%% file: main.tex
\documentclass{article}

\usepackage[final,nonatbib]{bdl_2018}

\usepackage[utf8]{inputenc} 
\usepackage[T1]{fontenc}    
\usepackage{booktabs,ragged2e,amsfonts,tabularx,hyperref,url,times,
courier,csquotes,dblfloatfix,setspace,ifthen,amsmath, amsthm, amssymb, wrapfig,graphicx,amsfonts,multirow,subfig,todonotes,algorithmic,times,algorithm}

\author{
  Ershad Banijamali$^{1,2}$ , Amir-Hossein Karimi$^1$ , Ali Ghodsi$^3$ 
   \\  
  $^1$School of Computer Science,  University of Waterloo \\
    $^2$Vector Institute \\
  $^3$Department of Statistics and Actuarial Science, University of Waterloo \\
  \small
  \texttt{ \{sbanijam,a6karimi,aghodsib@uwaterloo.ca\} } \\
}

\title{Deep Variational Sufficient Dimensionality Reduction}

\begin{document}

\maketitle

\begin{abstract}
We consider the problem of sufficient dimensionality reduction (SDR), where the high-dimensional observation is transformed to a low-dimensional sub-space in which the information of the observations regarding the label variable is preserved.  We propose DVSDR, a deep variational approach for sufficient dimensionality reduction. The deep structure in our model has  a bottleneck that represent the	 low-dimensional embedding of the data. We explain the SDR problem using graphical models and use the framework of variational autoencoders to maximize the lower bound of the log-likelihood of the joint distribution of the observation and label. We show that such a maximization problem can be interpreted as solving the SDR problem. DVSDR can be easily adopted to semi-supervised learning setting. In our experiment we show that DVSDR performs competitively on classification tasks while being able to generate novel data samples.
\end{abstract}

\input{Introduction}

\input{Model_description}

\input{Experiments}

\newpage
\bibliography{DVSDR_ref}
\bibliographystyle{abbrv}

\end{document}

%% file: Introduction.tex
\section{Introduction}
\label{sec:intro}
Dimensionality reduction is a long-standing problem in machine learning. The initial motivation behind dimensionality reduction was to visualize data and many unsupervised, supervised, and semi-supervised algorithms have been designed for this purpose. 
Recent advancements in the area of deep learning has pushed the boundaries of the state-of-the-art across a variety of fields including object classification \cite{szegedy2015going,krizhevsky2012imagenet,simonyan2014very}, speech recognition \cite{graves2013speech,hinton2012deep}, and sample synthesis \cite{goodfellow2014generative,reed2016generative,isola2016image}, among others. Casting classical algorithms in statistical machine learning within the framework of deep learning has been transformative for the aforementioned applications, e.g. \cite{chan2015pcanet}.  This inspired us to revisit the problem of sufficient dimensionality reduction and tackle it using the tools provided to us by deep learning.

Sufficient dimensionality reduction (SDR) \cite{globerson2003sufficient} is a technique that aims to find a low-dimensional representation of data that retains predictive information regarding the label (response) variable. The original paper of SDR presents a way of quantifying this information using information theoretic notions and introduces an iterative algorithm for extracting the features that maximizes this information. Although SDR methods are typically applied to continuous target variables, there exist methods based on distance covariance that can estimate the central subspace, the intersection of all dimensionality reduction subspaces, for discrete target variables \cite{sheng2016sufficient}. In this paper we consider the discrete target scenario.

%% file: Model_description.tex
\newcommand\independent{\protect\mathpalette{\protect\independenT}{\perp}}
\def\independenT#1#2{\mathrel{\rlap{$#1#2$}\mkern2mu{#1#2}}}

\section{Model description}
In this section we first overview SDR and then present a graphical model interpretation of SDR and  propose a deep model in the framework of variational autoencoder that approximates the posterior in the graphical model and learns the low-dimensional space efficiently.

\subsection{Sufficient Dimensionality Reduction}

Consider the frequently encountered goal of regression: predicting a future value for a univariate label $\mathbf{y}\in \mathbb{R}^D$ for a given observation $\mathbf{x} \in \mathbb{R}^p$. In large scale domains, traditional regression methods may require large amounts of training data to avoid overfitting. Therefore, there is a pertinent need for dimensionality reduction methods that replace the original covariate $\mathbf{x}$ with another variable $\mathbf{z} \in \mathbb{R}^d$ that retains most or all of the information and variation in $\mathbf{x}$. When $\mathbf{z}$ retains all the relevant information about $\mathbf{y}$, the dimensionality reduction is said to be \textit{sufficient}: 
$p(\mathbf{y}|\mathbf{z}(\mathbf{x})) = p(\mathbf{y}|\mathbf{x}).$

Alternatively, sufficient dimension reduction (SDR) techniques can be viewed as methods that find a low dimensional representation such that the remaining degrees of freedom become conditionally independent of the output values, i.e.:

\begin{equation}
	\mathbf{y} \independent \mathbf{x} ~ | ~ \mathbf{z}
    \label{eq:sdr_2}
\end{equation}

In other words, $\mathbf{z}(\mathbf{x})$ carries all the information of $\mathbf{x}$ needed to predict a future value for $\mathbf{y}$.


\subsection{SDR, a graphical model interpretation}
\label{sec:model}
\begin{figure*}[!h]
\vspace{-.2cm}
    \centering
    \subfloat[]{{\includegraphics[trim = -10mm 3mm -10mm 2mm,height=2.5cm]{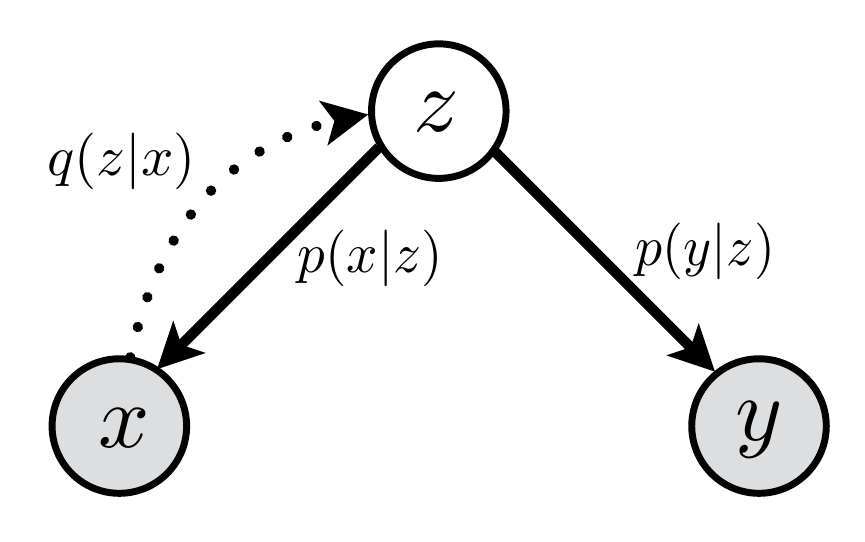}\label{fig:gm_2_1}}}
	\subfloat[]{{\includegraphics[trim = -10mm 0mm -10mm 0mm,height=2.5cm]{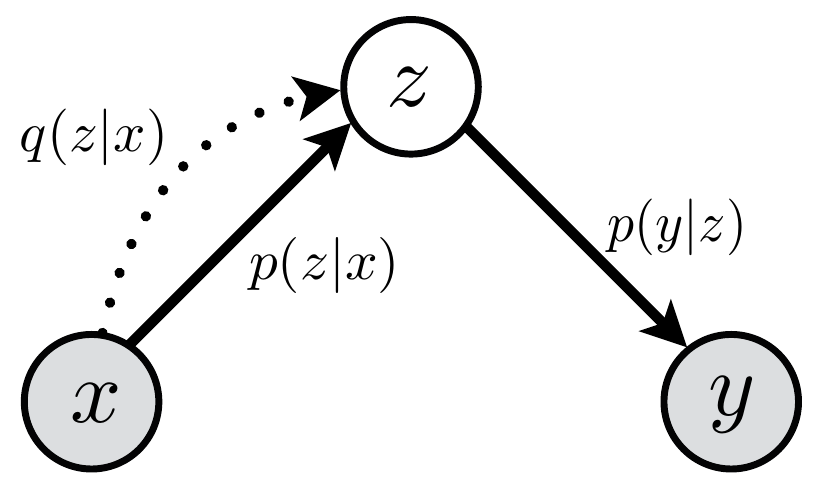} 	 \label{fig:gm_2_2}}} 
   \vspace{-.2cm}
    \caption{Candidate graphical models for sufficient dimensionality reduction } 
   \label{fig:gm_2}
\end{figure*}

The goal in SDR, as stated in above, is to discover a latent space that can be used for classification. Therefore, compared to the unsupervised dimensionality reduction algorithms, SDR  aims to find a latent variable with high predictive power. SDR problem can be explained by either  of the graphical models in Fig. \ref{fig:gm_2}. Although the objective of SDR might be derived from Fig. \ref{fig:gm_2_2}  more intuitively, we argue that the latent space that is learned in Fig. \ref{fig:gm_2_1}  carries more information about $\mathbf{x}$ and therefore is more generalizable. In fact, deriving $p(\mathbf{y}|\mathbf{x})$ from the joint distribution $p(\mathbf{x},\mathbf{y})$ yields a better representation in the latent space that is more robust against overfitting \cite{shu2017bottleneck}. This is especially true when we parameterize these conditional probabilities using neural networks with hundreds of thousands of parameters. Another benefit of the model in Fig. \ref{fig:gm_2_1} is that it can leverage unlabeled samples to build the latent space and therefore may be used for semi-supervised learning. Whereas the model in Fig. \ref{fig:gm_2_1} can only use  the labeled samples. 

In fact, we assume that $\mathbf{z}$ can not only generate high-dimensional observations, but can also generate the target variable. Therefore, our objective is to find a latent variable that keeps the information of $\mathbf{X}$ and can be used for classification. In the graphical model of Fig. \ref{fig:gm_2_2}, this is equivalent to maximizing the joint distribution of evidences, $p(\mathbf{X},\mathbf{Y})$ for all pairs of $(\mathbf{x}_i, \mathbf{y}_i)$. One way to maximize this likelihood is using the variational inference. Based on this graphical model the variational lower bound of the log-likelihood of the joint distribution can be written as:
\begin{equation}
\log p(\mathbf{x},\mathbf{y})  \geq  \mathbb{E}_{q(\mathbf{z}|\mathbf{x})} \big [ \log p(\mathbf{x}|\mathbf{z}) + \log p(\mathbf{y}|\mathbf{z})\big] - \text{KL} \big ( q(\mathbf{z}|\mathbf{x}) \parallel p(\mathbf{z})\big).
\label{eq:elbo}
\end{equation}
We assume that the prior distribution $p(\mathbf{z})= \mathcal{N}(\mathbf{0},\mathbf{I})$. Our goal is to maximize this lower bound. In this model the posterior approximating distribution $q(\mathbf{z}|\mathbf{x})$ plays the role of a probabilistic dimensionality reduction function, $\mathbf{z}(\mathbf{x}) \sim q(\mathbf{z}|\mathbf{x}) $.

\subsection{Deep Variational Learning}
Variational autoencoders (VAEs) \cite{vae2014,rezende2014stochastic} are deep models that can maximize the lower bounds of type Eq. \ref{eq:elbo} efficiently in large scale. In these models, the conditional distributions are parameterized by neural networks. An \textit{encoder } network parameter set $\phi$ models $q_{\phi}(\mathbf{z}|\mathbf{x})$, which is a transformation function that maps high-dimensional observation to the latent space. A \textit{decoder} with parameter set $\theta$ models $p_{\theta}(\mathbf{x}|\mathbf{z})$, which is a mapping from the latent space to observation space. And a \textit{classifier} with parameter set $\psi$ models $p_{\theta}(\mathbf{y}|\mathbf{z})$, which is a mapping from the latent space to the space of labels. Therefore the lower bound in \ref{eq:elbo}, denoted by $\mathcal{L}(\mathbf{x},\mathbf{y})$ can be rewritten as:

\begin{equation}
\begin{array}{ll}
\mathcal{L}_{\phi,\theta,\psi} (\mathbf{x},\mathbf{y}) & = \mathbb{E}_{q_{\phi}(\mathbf{z}|\mathbf{x})} \big [ \log p_{\theta}(\mathbf{x}|\mathbf{z}) \big ] + \mathbb{E}_{q_{\phi}(\mathbf{z}|\mathbf{x})}  \big [ \log p_{\psi}(\mathbf{y}|\mathbf{z})\big] - \text{KL} \big ( q_{\phi}(\mathbf{z}|\mathbf{x}) \parallel p(\mathbf{z})\big). 
\end{array}
\label{eq: obj_1}
\end{equation}

Our approach to solve the SDR problem is to maximize this lower bound using deep variational learning, and we call it deep variational SDR (DVSDR).  
%
By maximizing $\mathbb{E}_{q_{\phi}(\mathbf{z}|\mathbf{x})} \big [\log p_{\psi}(\mathbf{y}|\mathbf{z})\big]$ for each labeled sample, we build  a latent representation that has a high predictive power and this is exactly what SDR aims for. 



\subsection{Semi-supervised DVSDR}
In many cases in practical problems, the label information is scarce and therefore a semi-supervised learning method should be exploited. DVSDR can be easily adopted to a semi-supervised learning setting. Suppose $\mathbf{X}_L \subset \mathbf{X}$ is the set of all labeled samples for which we have the label set $\mathbf{Y}$, and $\mathbf{X}_U = \mathbf{X} \backslash \mathbf{X}_L$ is the set of unlabeled samples.  Since our latent variable is inferred purely based on our observation variable (and not label variable), we can split the objective function to two parts: $\mathcal{L}^{\ell}_{\phi,\theta,\psi} (\mathbf{x},\mathbf{y})$, which is equal to $\ref{eq: obj_1}$, for the labeled points and  $ \mathcal{L}^\mathit{u}_{\phi,\theta} (\mathbf{x}) $, for the unlabeled points, which has the following form:

\begin{equation}
\mathcal{L}^u_{\phi,\theta} (\mathbf{x}) = \mathbb{E}_{q_{\phi}(\mathbf{z}|\mathbf{x})} \big [ \log p_{\theta}(\mathbf{x}|\mathbf{z}) \big ]  - \text{KL} \big ( q_{\phi}(\mathbf{z}|\mathbf{x}) \parallel p(\mathbf{z})\big). 
\end{equation}
 
  $\mathcal{L}^u_{\phi,\theta} (\mathbf{x}) $ has the same form as a regular VAE. Therefore the overall objective of the semi-supervised model is:
\begin{equation}
 \max \limits_{\phi,\theta,\psi} \sum_{\mathbf{x}\in \mathbf{X}^{\ell}}\mathcal{L}^{\ell}_{\phi,\theta,\psi} (\mathbf{x},\mathbf{y}) + \sum_{\mathbf{x}\in \mathbf{X}^{\mathit{u}}} \mathcal{L}^\mathit{u}_{\phi,\theta} (\mathbf{x}) .
\end{equation}
In fact, in the semi-supervised setting, the low-dimensional latent variable for unlabeled samples is used to only reconstruct the observations, but for the labeled samples it is used to reconstruct the samples and predict the labels.

%% file: Experiments.tex
\newcommand{\oo}[2]{#1{\tiny/#2}}
\newcolumntype{L}[1]{>{\raggedright\let\newline\\\arraybackslash\hspace{0pt}}m{#1}}
\newcolumntype{C}[1]{>{\centering\let\newline\\\arraybackslash\hspace{0pt}}m{#1}}
\newcolumntype{R}[1]{>{\raggedleft\let\newline\\\arraybackslash\hspace{0pt}}m{#1}}

\section{Experiment Results}
\label{sec:exp}
We evaluate the performance of DVSDR in two different tasks: Classification and new sample generation. 

%
%
%
\subsection{Classification}
We compare the classification performance of the proposed model with Adversarial Autoencoder (AAE) \cite{makhzani2015adversarial} and semi-supervised learning with deep generative models \cite{kingma2014semi} that are both deep generative models with bottleneck. 

\begin{table}[!h]
\centering
\begin{tabular}{l|| c | c }
                          & MNIST (All labeled samples)             &  MNIST (1000 labeled samples)              \\ \hline
 VAE (M1+M2) 		&  0.96 $\pm$ 0.03 &  2.40 $\pm$ 0.02 \\ \hline
  AAE 		&  0.86 $\pm$ 0.03 & \textbf{ 1.60} $\pm$\textbf{ 0.08}  \\ \hline
  DVSDR     &  \textbf{0.80 }$\pm$\textbf{ 0.04}& 2.1 $\pm$ 0.06  \\
\end{tabular}
\vspace{.5cm}
\caption{Classification Error rate}
\end{table}

As we can see our model outperforms the other two models when we use all the label information, but when only $1000$ labeled samples are used, AAE performs better. This is because AAE directly impose a supervised loss on the latent space by matching the distribution of the latent space with a categorical distribution.

\subsection{Generation}
Sample generation is a bi-product of the DVSDR algorithm. Using the structure of DVSDR not only can we preserve the information in the observation set while reconstructing the input samples, but also we can  generate novel data by sampling from the prior $p(\mathbf{z})$ and feeding it to the decoder. 

\begin{figure*}[!h]
    \centering
    \subfloat[2-dimensional latent space]{{\includegraphics[trim = 0mm 0mm 0mm 0mm,width=12cm]{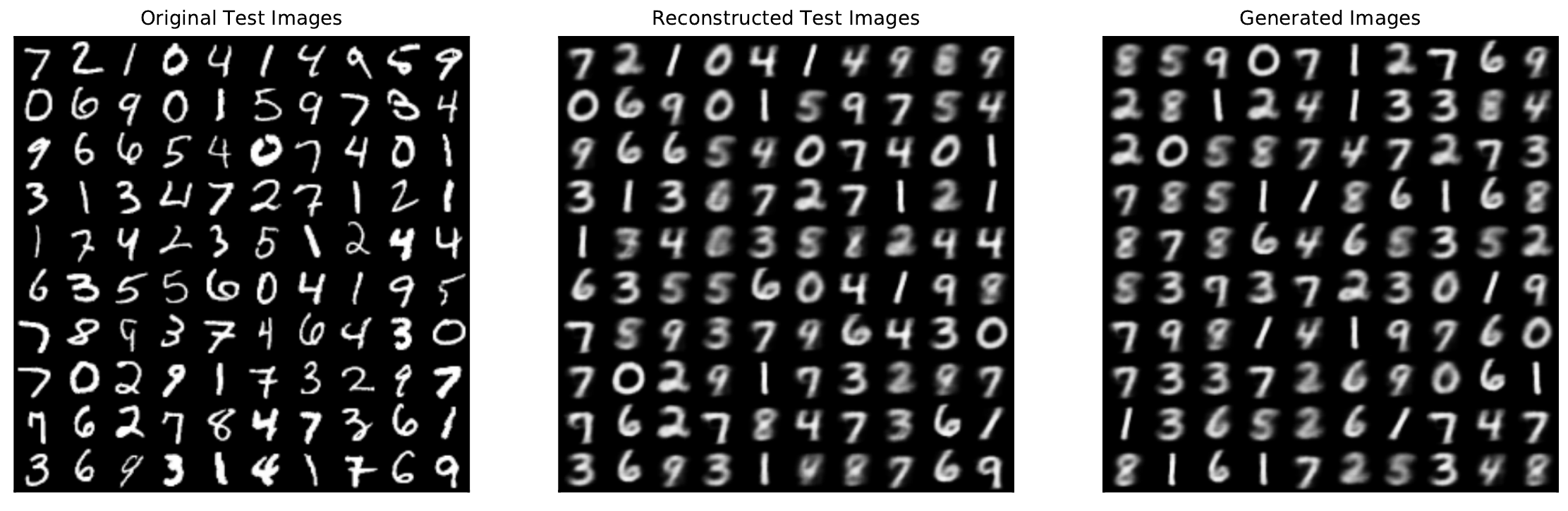}\label{fig:MNIST_1}}} \\
    \subfloat[15-dimensional latent space]{{\includegraphics[trim = 0mm 0mm 0mm 0mm,width=8cm]{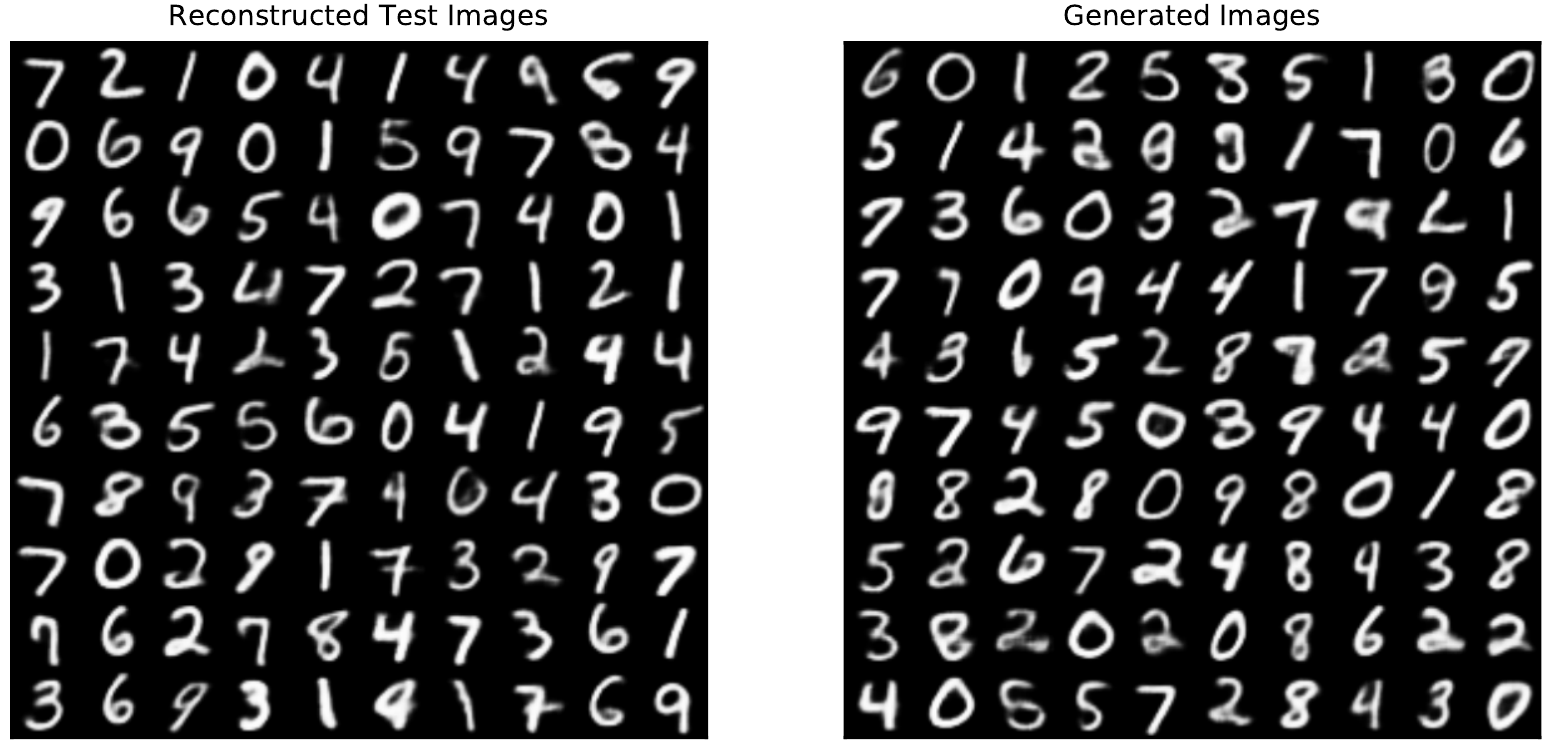}	 \label{fig:MNIST_2}}}
	\subfloat[]{{\includegraphics[trim = -10mm 0mm 0mm 0mm,width=4.2cm]{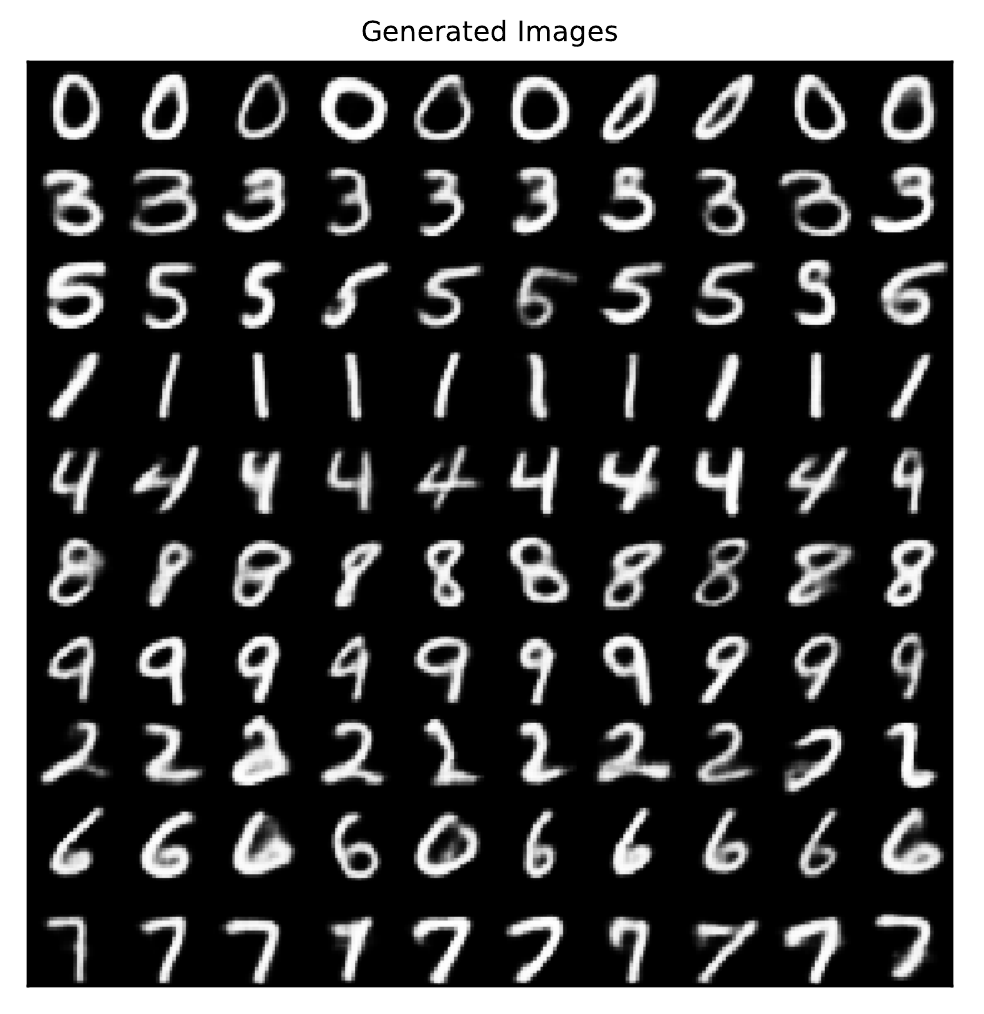}	 \label{fig:MNIST_3}}} 
    \caption{(a,b) Reconstructed and generated images using DVSDR with 2-D and 15-D latent space. (c) Fitting a mixture of Gaussian with 10 components on the latent space and sampling from each component } 
   \label{fig:MNIST}
\end{figure*}

\begin{figure*}[!h]
    \centering
 \includegraphics[trim = 0mm 0mm 0mm 0mm,width=14cm]{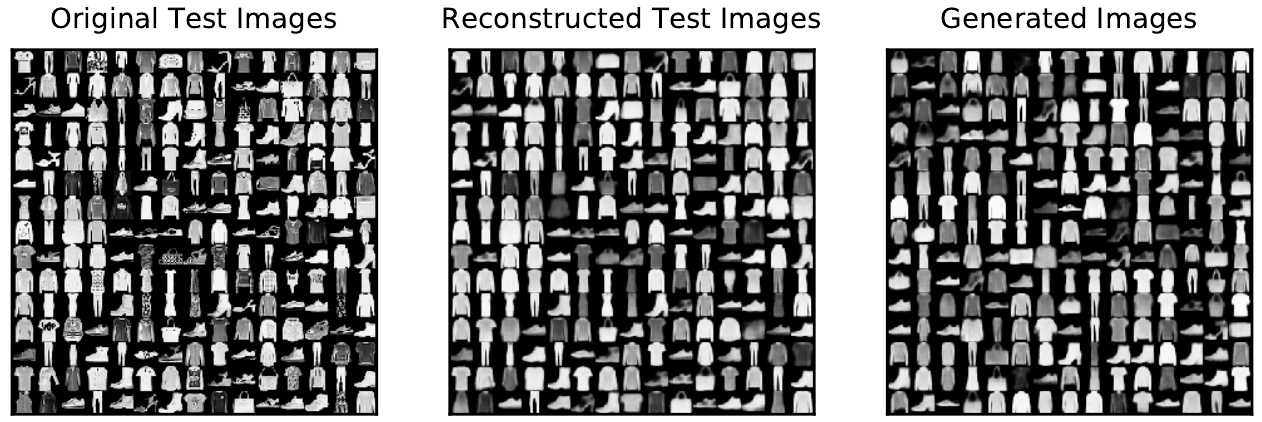}
    \caption{Fashion MNIST. 10-D latent space} 
   \label{fig:FMNIST}
\end{figure*}

Fig. \ref{fig:MNIST} and \ref{fig:FMNIST} show the results of sample reconstruction/generation for MNIST and Fashion MNIST datasets. For MNIST we use DVSDR with a  latent space of dimensionality  2 and 15, and for Fashion MNIST dimensionality of 10.  We can generate high quality images using the proposed model. When we fit a mixture of Gaussian over the latent space of the model for the MNIST dataset, we can generate samples from different classes, this is because the representations of the points in each class in the latent space are very well separated.

